\documentclass[10pt,twocolumn,letterpaper]{article}

 \usepackage{cvpr}              
\usepackage{multirow}
\usepackage{colortbl}
\usepackage{animate}

\usepackage[dvipsnames]{xcolor}

\definecolor{cvprblue}{rgb}{0.21,0.49,0.74}
\usepackage[pagebackref,breaklinks,colorlinks,citecolor=cvprblue]{hyperref}

\def\paperID{***} 
\def\confName{CVPR}
\def\confYear{2024}

\title{\vspace{-8mm}Diffusion-based Light Field Synthesis}

\author{Ruisheng Gao\thanks{\xspace\xspace Author contributed equally} \quad\quad Yutong Liu{\footnotemark[1]} \quad\quad Zeyu Xiao \quad\quad Zhiwei Xiong \\ 
University of Science and Technology of China \\
}

\begin{document}
\maketitle
\begin{abstract}
Light fields (LFs), conducive to comprehensive scene radiance recorded across angular dimensions, find wide applications in 3D reconstruction, virtual reality, and computational photography.
However, the LF acquisition is inevitably time-consuming and resource-intensive due to the mainstream acquisition strategy involving manual capture or laborious software synthesis.
Given such a challenge, we introduce LFdiff, a straightforward yet effective diffusion-based generative framework tailored for LF synthesis, which adopts only a single RGB image as input.
LFdiff leverages disparity estimated by a monocular depth estimation network and incorporates two distinctive components: a novel condition scheme and a noise estimation network tailored for LF data.
Specifically, we design a position-aware warping condition scheme, enhancing inter-view geometry learning via a robust conditional signal.
We then propose DistgUnet, a disentanglement-based noise estimation network, to harness comprehensive LF representations.
Extensive experiments demonstrate that LFdiff excels in synthesizing visually pleasing and disparity-controllable light fields with enhanced generalization capability.
Additionally, comprehensive results affirm the broad applicability of the generated LF data, spanning applications like LF super-resolution and refocusing.

\end{abstract}

\vspace{-3mm}
\section{Introduction}
\label{sec:intro}

Light field (LF) imaging enables the record of both intensities and directions of light rays in free space, providing a richer description of scene radiance than conventional 2D imaging techniques.
In particular, 4D LF plays an important role in its appealing applications in computer vision such as depth sensing \cite{wang2015occlusion}, post-capture refocusing \cite{ng2005light,bishop2011light}, reflectance estimation \cite{wang2016svbrdf} and salient object detection \cite{sheng2016relative}.
These applications benefit greatly from data-driven deep-learning approaches that leverage large-scale LF datasets~\cite{wang2016svbrdf,wang2022disentangling,wang2018selective,zhang2020lfnet,tsai2020attention,peng2020zero,peng2018unsupervised,liu2023light,xiao2023space,xiao2023toward,chao2023lfsrdiff}.

To acquire LFs, one straightforward approach involves manual capture using an LF camera, such as Lytro \cite{lytro}.
Besides, software rendering (\textit{e.g.}, Blender \cite{blender}) offers controllable camera parameters and scene contexts to synthesize desired data.
However, both of these methods require either time-consuming post-processing of raw data or labour-intensive manual design, presenting challenges such as data shortage and inconvenient acquisition for various applications (\textit{e.g.}, LF super-resolution, post-capture refocusing).

In response to these challenges, synthesizing LFs from single RGB images \cite{srinivasan2017learning,ivan2019synthesizing,li2020synthesizing,bak2023light} provide practical solutions to obtain sufficient LFs.
A fundamental approach is to warp input RGB images utilizing the estimated disparity, which suffers from the intrinsic and inevitable degradation on non-Lambertian regions (\textit{e.g.}, occlusions).
The prevalent approach utilizes deep neural networks (DNNs), leveraging monocular depth as an implicit prior to mapping single RGB images to ground truth LFs \cite{bak2023light, li2020synthesizing, ivan2019synthesizing, srinivasan2017learning}.
While these methods aim to mitigate non-Lambertian effects caused by the warping operation, implicitly predicting ambiguous non-Lambertian regions from local adjacent features can introduce artifacts and biases.
Furthermore, the disparity ranges of LFs synthesized by these methods heavily rely on training data, limiting their generalization across different geometry patterns.
On the other hand, LF synthesis from single images is an ill-posed problem.
For the same central view, multiple potential LF targets with different disparity ranges exist.
Unfortunately, the above depth-based methods lack the flexibility to synthesize LFs with controllable geometric.

Recently, the advanced diffusion models (DMs) demonstrate superior generative capability and remarkable performance, bringing novel paradigms~\cite{li2022srdiff,chen2023diffusiondet,ho2022imagen,blattmann2023align,luo2021diffusion,melas2023pc2} to the computational imaging tasks.
Composing of forward and backward processes, DM is able to model complex distribution for various modalities and achieve high quality generation with realistic details (\textit{e.g.}, images \cite{li2022srdiff,chen2023diffusiondet}, videos \cite{ho2022imagen,blattmann2023align} and point clouds \cite{luo2021diffusion,melas2023pc2}).
By further introduce the conditional signal, DMs allow for accurate and flexible controls for the generation process \cite{rombach2022high,zhang2023adding}. 
Considering the aforementioned merits, DMs serve as a compatible candidate to synthesis LFs from single images.

However, the complex spatial-angular pattern within 4D LFs poses challenges to adapting existing DMs to to synthesize LFs from single images.
On the one hand, to synthesize angular-correct and controllable LFs from single images, the condition signal needs to be designed to fully use the appearance information within single images while enabling geometry-aware guidance.
On the other hand, existing noise estimation networks are mostly designed for images and videos (2D or 3D Unet architecture), which struggle in capturing intra-inter view correlations within LFs.

Concerning the above issues, we propose LFdiff, a diffusion-based conditional generation framework tailored for LF synthesis from single images.
Given the intricacies of the 4D LF distribution, we design a position-aware warping scheme to provide a robust initial estimate of the 4D LF pattern, explicitly incorporating spatial-angular information into the condition signal.
Specifically, we use a pre-trained monocular depth estimation network to obtain the inverse monocular depth of the input single RGB image.
By rescaling the estimated inverse depth to disparities with varied ranges, we attain the capability to synthesize LFs with controllable disparaties.
Then we use the warp operation to create a coarse estimate of the target LF, which is further concatenated to a positional encoding to form the condition signal.
In addition, we introduce the disentangle mechanism \cite{wang2022disentangling} into the noise estimation network, resulting in DistgUnet.
Compared to the vanilla Unet, DistgUnet well leverages multiple LF representations from macro-pixel inputs, improving the overall generation quality.
Experimental results verify the ability of our framework to generate both angular-correct and visually pleasing LFs with controllable disparity ranges, further boosting various applications such as LF super-resolution and refocusing.
An example of the generated result is shown in Fig.~\ref{fig:teaser}.

The contributions of this paper are summarized as follows.
(1) We propose LFdiff, the first diffusion-based LF synthesis framework,
which includes two effective designs: a position-aware warping condition scheme for angular-aware guidance and a disentangling noise estimation network for enhanced spatial-angular expression.
(2) Extensive experiments demonstrate that LFdiff is able to generate LFs with accurate angular patterns, achieving superior visual-fidelity quality on central-view conditioned LF synthesis.
(3) LF synthesis from single images results verify the controllable generation as well as cross-domain generalization capability of LFdiff.
(4) Various applications such as LF super-resolution and refocusing validate our framework's broad applicability.

\section{Related Work}

\label{sec:related_work}

\noindent \textbf{Novel view synthesis from single images.}
Prior works synthesize novel views either from multiple views \cite{aliev2020neural,hedman2018deep,novotny2019perspectivenet,cao2022fwd}, or only from single images \cite{wiles2020synsin,li2022infinitenature,lin2023vision}.
The latter setting poses additional challenges for the lack of scene geometry priors.
Wiles \textit{et al.} \cite{wiles2020synsin} target on indoor/outdoor scene synthesis. 
They render a point cloud from estimated depth maps and content features, followed by a refinement network to inpaint unseen regions. 
InfiniteNature \cite{li2022infinitenature} renders novel views for natural scenes along a camera trajectory via a render-refine-repeat process in a self-supervised manner.
Besides, some neural radiance fields (NeRFs) based methods can render scene volume from a single view while requiring training for an individual scene.
Lin \textit{et al.} \cite{lin2023vision} focus on object-level view synthesis and design a transformer \cite{dosovitskiy2020image} inspired architecture to improve 3D feature expressiveness for the subsequent NeRF \cite{mildenhall2021nerf} rendering.
In contrast, we synthesize LFs from single images, demanding more accurate inter-view geometry than 3D synthesis without individual training for each scene.

\noindent \textbf{Light field synthesis from single images.}
As a pioneer work, Srinivasan \textit{et al.} \cite{srinivasan2017learning} breaks this task into two sub-tasks: monocular depth estimation and LF synthesis and learns each subtask using convolution neural networks (CNN).
Ivan \textit{et al.} \cite{ivan2019synthesizing} utilize the appearance flow as the geometry representation to preserve the spatial-angular consistency between views.
Li \textit{et al.} \cite{li2020synthesizing} extend the multi-plane representation and use a parallel CNN to deal with visible and occluded regions, respectively, showing improved synthesis quality.
Bak \textit{et al.} \cite{bak2023light} propose an improved variable layered depth image for scene representation, producing visually clearer results in fewer inference times.
Apart from the above works, LF synthesis from a monocular video \cite{govindarajan2022synthesizing} or coded view \cite{marwah2013compressive,vargas2021time,mizuno2022acquiring} also provide insights from temporal information utilization and hardware-level compressive imaging, respectively.
Unlike above frameworks, we propose a diffusion-based generative framework to synthesize disparity controllable LFs with improved visual results and enhanced generalization ability.

\noindent \textbf{Diffusion based view synthesis.}
Diffusion models have demonstrated significant improvement in novel view synthesis conditioned on geometric priors such as poses \cite{watson2022novel,tseng2023consistent}, 3D feature volumes \cite{chan2023generative} or semantic priors \cite{tseng2023consistent}.
3DiM \cite{watson2022novel} leverages the pose between two images and proposes a 2D diffusion model to generate novel views in an auto-regressive manner.
Chan \textit{et al.}\cite{chan2023generative} unproject multi-view features into a feature volume to regress the density and content feature, which serve as the condition signal.
NeRDi \cite{gu2023nerfdiff} utilized a language-guided diffusion prior for multi-view synthesis, which links image semantics to the appearance reconstruction.
To explicitly capture multi-view geometry in LFs, we propose a position-aware warping scheme to provide a coarse LF estimate as the condition signal.

\begin{figure*}[ht!]
    \centering
    \scalebox{1}
    {\includegraphics[width=\linewidth]
    {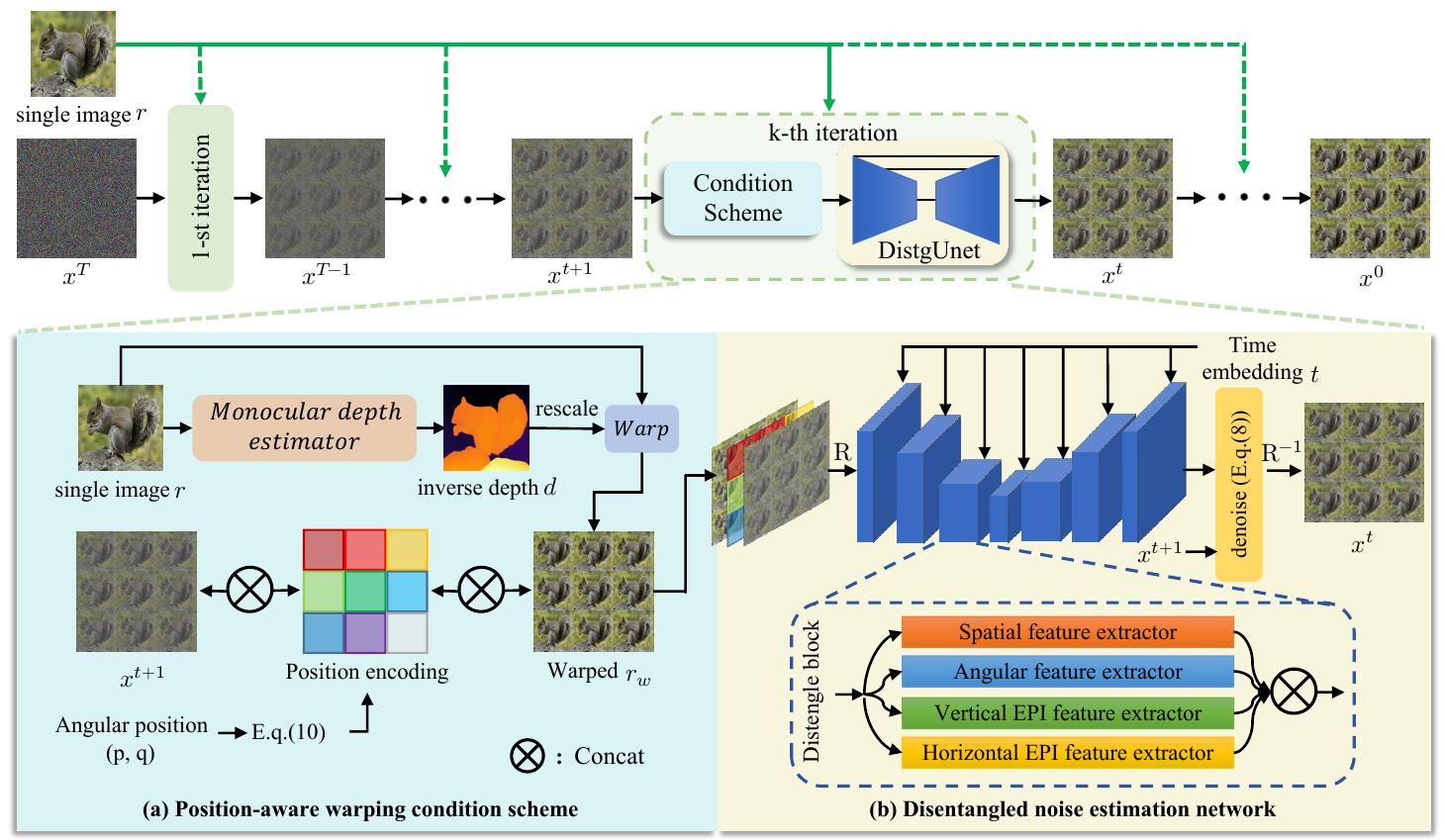}}
    \caption{The overall framework of LFdiff. After iterating for $T$ timesteps, random Gaussian noise $x_{T}$ is denoised into a high-quality LF $x_{0}$. We provide the details of the position-aware warping condition scheme in (a) and depict the disentangled noise estimation network in (b). $\mathrm{R}$ and $\mathrm{R^{-1}}$ denotes SAI to macro-pixel reshape and the inverse reshape, respectively. In the training stage, we use the ground-truth disparity instead of estimated invert depth to warp the LF central view, which is omitted in (a) for simplicity.}
    \label{Fig:main}
\end{figure*}

\section{Preliminaries}
\label{sec:3}
We first provide a concise introduction to the learning objective of conditional DM in the context of LF inputs.
Given a LF $x\in \mathbb{R} ^{U\times V\times H\times W\times C} $ with spatial resolution $H\times W$ and angular resolution $U\times V$, we can represent it in sub-aperture images (SAIs) $x_0\in \mathbb{R} ^{UH\times VW\times C} $, which lies in the target distribution for learning.
The Denoising Diffusion Probabilistic Model (DDPM) includes a forward process which repeatedly adds noise to the target, transforming $x_0$ to a normal Gaussian noise $x_T$ in $T$ timesteps.
Each forward step is given by
\begin{equation}
    q(x_{t}\mid x_{t-1} ) = \mathcal{N}(x_{t};\sqrt{1-\beta_{t}}x_{t-1}, \beta_{t}\mathrm {I}),
\end{equation}
where $\left \{ \beta_{t} \right \} _{t=1}^{T} $ are predefined as the noise schedule, $\mathrm{I}$ refers to the identity matrix. Using the reparameterization trick\cite{kingma2013auto}, we can obtain $x_{t}$ in one step as
\begin{equation}
    q(x_{t}\mid x_{0} ) = \mathcal{N}(x_{t};\sqrt{\bar{\alpha} _{t} }x_{0}, (1-\bar{\alpha} _{t})\mathrm {I}),
\end{equation}
where $\alpha_{t} = 1-\beta_{t}$, $\bar{\alpha}_{t}= {\textstyle \prod_{i=1}^{t}} \alpha_{i}$.

The reverse process starts from a randomly sampled normal Gaussian noise $x_{T}$ and aims to gradually denoise it to a high quality output $x_{0}$. 
To approximate the true posterior $q(x_{t-1}\mid x_{t})$ in each denoising step, parameterized gaussian transitions $p_{\theta} (x_{t-1}\mid x_{t})=\mathcal{N}(x_{t-1};\mu _{\theta }(x_t,t), \Sigma _{\theta }(x_t,t)) $ are assumed as \cite{ho2020denoising} and we use trainable networks to learn the mean $\mu _{\theta }(x_t,t)$ and the variance $\Sigma _{\theta }(x_t,t)$. In the DDPM setting, the variances are set to a fixed value $\sigma _{t}^{2}\mathrm {I} $, and by optimizing the variational lower bound on negative log-likelihood
\begin{equation}
L_{\theta} \le  \mathbb{E}_{q(x_{0})}[-\mathrm {log}p_{\theta}(x_{0})] ,
\end{equation}
we can obtain a simpler training objective through further simplifications\cite{ho2020denoising}
\begin{equation}
L_{simple}=\mathbb{E}_{x_{0},\varepsilon\sim \mathcal{N}(0,\mathrm {I}   ),t  }[\left \| \varepsilon-\varepsilon_{\theta}(x_{t},t) \right \|_{2}^{2}  ] ,
\end{equation}
where $\varepsilon_{\theta}$ is the noise estimation network (\textit{e.g.}, Unet) and $t$ is uniformly sampled from $\left \{  1...T\right \} $.
Furthermore, the generation process can be guided when additional conditions $c$ are provided.
In this way, the training objective becomes
\begin{equation}
L_{simple}=\mathbb{E}_{x_{0},\varepsilon\sim \mathcal{N}(0,\mathrm {I}   ),t,c  }[\left \| \varepsilon-\varepsilon_{\theta}(x_{t},t,c) \right \|_{2}^{2}  ] .
\end{equation}

In this paper, we provide special designs on the condition scheme and the noise estimation network for synthesis LFs, which are elaborated in the following section.

\begin{figure}[ht!]
    \centering
    \scalebox{1}
    {\includegraphics[width=\linewidth]
    {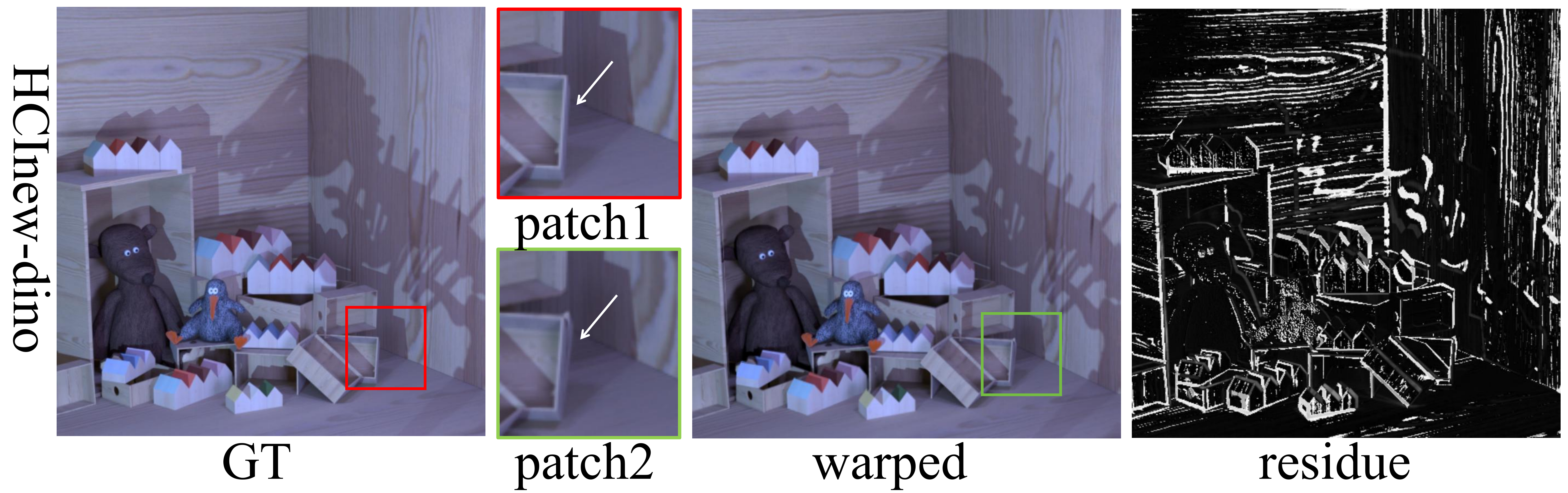}}
    \caption{Although the warp operation introduces occlusion artifacts (compare between patch1 and patch2) and spatial misalignment (see residue map), it provides an initial LF pattern for guidance. We use residue between GT and warped view for clarity.}
    \label{Fig:main2}
\end{figure}

\section{Method}
\label{sec:method}
The overall framework of LFdiff is shown in Fig.~\ref{Fig:main}. Given a single RGB image $r\in \mathbb{R} ^{H\times W\times 3} $, we first utilize a pre-trained monocular depth estimator \cite{ranftl2020towards} to obtain the normalized inverse depth $d\in \mathbb{R} ^{H\times W}$. After rescaling $d$ to desired disparity range [$d_{min}$, $d_{max}$], we can get the condition of our framework $c\in \mathbb{R} ^{UH\times VW\times 4}$ by the proposed position-aware warping condition scheme $\tau $

\begin{equation}
    c = \tau(r,d).
\end{equation}

Then, through an iterative denoising (sampling) process (\textit{e.g.}, DDPM \cite{ho2020denoising}, DDIM \cite{song2020denoising}), we can get the synthesised LF $o\in \mathbb{R} ^{UH\times VW\times 3}$ from a randomly sampled noise $x_{T}\sim \mathcal{N}(0,\mathrm {I}) $.
At time step $t$, $x_{t+1}$ and condition $c$ are concatenated along the channel dimension and reshape into the macro-pixel form, serving as the input of noise estimation network $\varepsilon _{\theta } $ as
\begin{equation}
   n_{est} =\mathrm {R^{-1}}(\varepsilon _{\theta } ( \mathrm {R} (x_{t+1}\otimes  c) )),
\end{equation}

where $n_{est}$ is the estimated noise, $\mathrm {R}$ and $\mathrm {R^{-1}}$ refer to the SAI to macro-pixel reshape operation and its reverse, respectively. $\otimes$ denotes the channel-wise concatenation. Then we can get $x_{t}$ through one-step denoising (DDPM for example) as
\begin{equation}
\label{Eq:8}
    x_{t}=\frac{1}{\sqrt{\alpha _{t+1} } } (x_{t+1}-\frac{1-\alpha _{t+1}}{\sqrt{1-\bar{\alpha}_{t+1}}}n_{est })+\sigma _{t+1}z,
\end{equation}
where $z\sim \mathcal{N}(0,\mathrm {I})$ when $t>0$, otherwise $z=0$. After $T$ iterations, we obtain the generated SAI form LF $x_{0}$.

\subsection{Position-aware Warping Condition Scheme}
\label{sec:3.2}

By incorporating condition signals into DMs, a controlled generation process can be achieved which flourishes diverse appealing applications (\textit{e.g.}, text-to-image/video generation \cite{rombach2022high,wu2023tune,karras2023dreampose}, super-resolution \cite{saharia2022image}, molecule synthesis\cite{hoogeboom2022equivariant}).
Specifically, prior works introduce the condition signal in different ways on different tasks, including direct concatenation \cite{saharia2022image}, vector embedding \cite{ho2022classifier}, learned cross-attention \cite{rombach2022high} and so on. 

In synthesizing light fields with complex spatial-angular distributions, a meticulously designed condition signal is essential to provide effective guidance.
On the one hand, LFs are spatial-angular intertwined data.
DMs can hardly distinguish each dimension to produce LFs with correct geometry and visually pleasing details. 
Thus, it is challenging for the DM to directly model 4D LF pattern solely from the guidance of the input single image. 
On the other hand, synthesizing LF from single images is an ill-posed problem since the disparity of generated LF varies according to different camera parameters, such as baseline and focus length.
Therefore, the condition signal is expected to be angular-aware, which allows for flexible control over variant geometry requirements.

Our solution to the above concerns is to explicitly utilize the estimated disparity to warp the input RGB image, resulting in a coarse estimate of the LF goal. As shown in Fig.~\ref{Fig:main}(a), given a single image $r$ and the rescaled inverse depth $d$, the warp operation acts as
\begin{equation}
    r_{w}^{i}(s,t) = r(s+(p_{c}-p_{i})\cdot d, t+(q_{c}-q_{i})\cdot d),
    \label{E.q.warp}
\end{equation}
where $r_{w}^{i}$ represents the $i$-th warped view, $s$ and $t$ denote spatial coordinate, $(p_{c},q_{c})$ and $(p_{i},q_{i})$ are the 2D angular-coordinate tuple of the central view and the $i$-th view, respectively.
Despite the warp operation introducing occlusion artifacts and spatial misalignment, as shown in Fig.~\ref{Fig:main2}, the warped LF serves as a reliable guidance of initial LF pattern which contains abundant spatial-angular information, while other possible conditions (\textit{e.g.}, single image only, depth embedding) are hard to represent the 4D LF characteristics (see Fig.~\ref{Fig:ablation}(a) and (b)).
In addition, benefiting from the angular-aware nature of the warp operation, the geometry of the warped result is controllable by rescaling the estimated disparity to different ranges.

Due to the unique nature of the angular position in light fields, embedding positional information into the condition signal is crucial.
Inspired by \cite{dosovitskiy2020image}, we introduce view-level 2D position encoding to assign each view a specific position, surpassing ambiguity in angular patterns across diverse generated light field samples (see Fig.~\ref{Fig:ablation}(c)).
Concretely, the position encoding for angular coordinate $(p, q)$ is defined as
\begin{equation}
	\begin{split}
			PE_{2i}(p,q)=sin(p/10000^{\frac{2i}{dim}} )+sin(q/10000^{\frac{2i}{dim}} ), \\
					PE_{2i+1}(p,q)=cos(p/10000^{\frac{2i+1}{dim}} )+cos(q/10000^{\frac{2i+1}{dim}} ),
	\end{split}
\end{equation}

where $dim$ denotes the encoding dimension, which is set to 16 in our experiments. 
Thus, $PE\in \mathbb {R }^{UH\times VW\times dim}$ provides the same position information within the same view while distinguishing positions across different views.
Given such designs, we can construct the condition signal by concatenating the warped results and the view-level position encoding along the channel dimension as

\begin{equation}
    c=r_{w} \otimes PE.
\end{equation}

\subsection{Disentangled Noise Estimation Network}
\label{sec:3.3}
After constructing the condition signal $c$, we can estimate the noise to be removed at timestep $t$ as

\begin{equation}
     \bar{\varepsilon} =\varepsilon _{\theta} (x_{t},t,c ),
\end{equation}
where $\bar{\varepsilon}$ is the estimated noise.
A typical choice of the noise estimation network is the Unet architecture \cite{ho2020denoising}, which is a multi-scale network with spatial downsampling-upsampling layers.
However, these 2D convolution operations cannot capture the complete representations along the spatial-angular dimensions for macro-pixel form LF inputs.

To incorporate more comprehensive LF representations, we resort to the disentangling mechanism \cite{wang2022disentangling} which disentangles the 4D LF into four different 2D subspaces: spatial space, angular space, horizontal and vertical epipolar plane image (EPI) spaces. 
Specifically, for different 2D subspaces, the mechanism employs specific 2D convolution structures. These structures are tailored to extract disentangling features from the intertwined 4D LF, thereby capturing and embedding domain-specific information.

We incorporate the disentangling mechanism into the vanilla 2D Unet architecture, resulting in the disentangled noise estimation network, dubbed DistgUnet.
As can be seen in Fig.~\ref{Fig:main}(b), we concatenate the condition signal and noisy inputs along the channel dimension to serve as inputs of the DistgUet.
Then the inputs are reshaped from SAIs to macro-pixel form, which further go through several stacked disentangle blocks in multiple scales.
In this way, multi-scale LF representations can be captured, which benefits the noise estimation process, therefore improving the overall generation quality. 
Please refer to the supplementary material for more details.

\begin{table*}[!t]
\centering
\caption{Quantitative results (PSNR $\uparrow$  / SSIM $\uparrow$  / LPIPS $\downarrow$ ) on the central view LF synthesis task. The best results are marked in \textbf{bold}. }
\label{Tab:syn}
\scalebox{0.85}{
\begin{tabular}{@{}l|cc|ccc@{}}
\toprule
& \multicolumn{2}{c|}{In distribution} & \multicolumn{2}{c}{Out of distribution} 
\\ \midrule

Method & \multicolumn{1}{c}{HCI-new} & \multicolumn{1}{c|}{UrbanLF-syn} 
& \multicolumn{1}{c}{HCI-old} & \multicolumn{1}{c}{STFGantry} \\ \hline

\multirow{1}{*}{Warp}  & 29.438/0.8931/0.048 &  31.866/0.9430/0.043 
& 30.943/0.8534/0.045 & 22.340/0.7482/0.079 \\

\multirow{1}{*}{Srinivasan \textit{et.al} \cite{srinivasan2017learning}}  & 27.175/0.7678/0.061  & 30.565/0.9308/0.035 & 29.608/0.7946/0.053 & 20.874/0.6746/0.086 \\

\multirow{1}{*}{Li \textit{et.al} \cite{li2020synthesizing}}  & 27.202/0.7782/0.060 &  31.334/0.9368/0.024 & 31.673/0.8806/0.046 & 21.651/0.7133/0.072 \\

\multirow{1}{*}{Bak \textit{et.al} \cite{bak2023light}}  & 27.930/0.7955/0.066  & 32.141/0.9380/0.027 & 31.932/0.8658/0.047 & 21.747/0.7021/0.076\\

\multirow{1}{*}{LFdiff (Ours)}  & \textbf{30.665/0.9135/0.025}
 &  \textbf{35.987/0.9712/0.016}  & \textbf{33.600/0.9207/0.023} & \textbf{24.264/0.7850/0.068} \\  
\bottomrule
\end{tabular}
}

\end{table*}

\begin{figure*}[h!]
    \centering
    \includegraphics[width=1\linewidth]{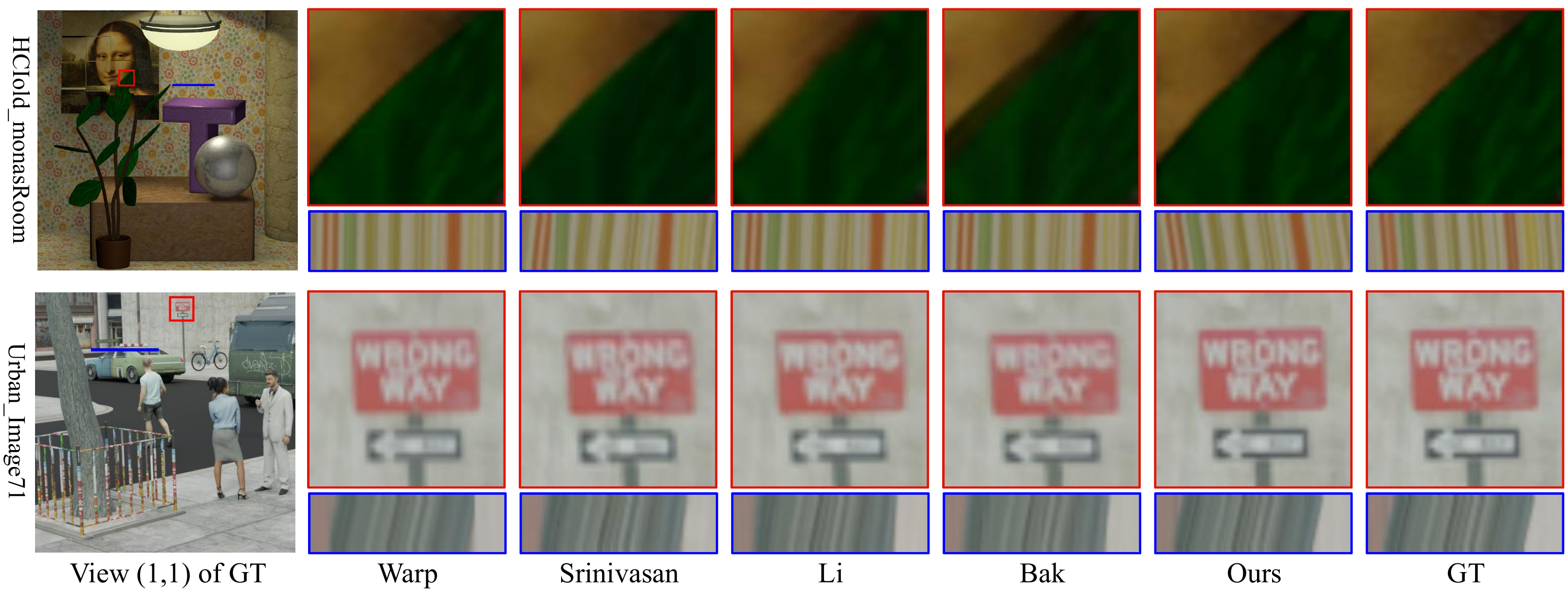}
    \caption{Qualitative comparisons including the SAIs and EPIs of synthesized LFs from central view through different methods along with the ground truth (view coordinates: (1, 1)). Zoom in for a better visual experience.}
    \label{Fig:synthesis_visual}
\end{figure*}

\begin{table}[h!]
\centering
\caption{Quantitative results (NIQE $\downarrow$  / BRISQUE $\downarrow$) on LF synthesis from single images. We use a subset of DIV2K and NYUV2 dataset to evaluate the perceptual quality through no-reference metrics. The best results are marked in \textbf{bold}. }
\label{Tab:syn2}
\scalebox{0.9}{

\begin{tabular}{@{}l|cc@{}}
\toprule

Method & \multicolumn{1}{c}{DIV2K} & \multicolumn{1}{c}{NYUV2} \\ \hline

\multirow{1}{*}{Warp}  & 4.14/17.98 &  5.00/37.77 \\

\multirow{1}{*}{Srinivasan \textit{et.al} \cite{srinivasan2017learning}}  & 4.20/17.59  & 4.97/36.70 \\

\multirow{1}{*}{Li \textit{et.al} \cite{li2020synthesizing}}  & 4.15/17.20 & 4.77/36.93  \\

\multirow{1}{*}{Bak \textit{et.al} \cite{bak2023light}}  &  4.34/19.24 & 5.02/38.23 \\

\multirow{1}{*}{LFdiff (Ours)}  & \textbf{4.06/13.72}
 &  \textbf{4.31/34.23}  \\  
\bottomrule
\end{tabular}
}
\end{table}

\begin{figure}[h!]
    \centering
    \includegraphics[width=1\linewidth]{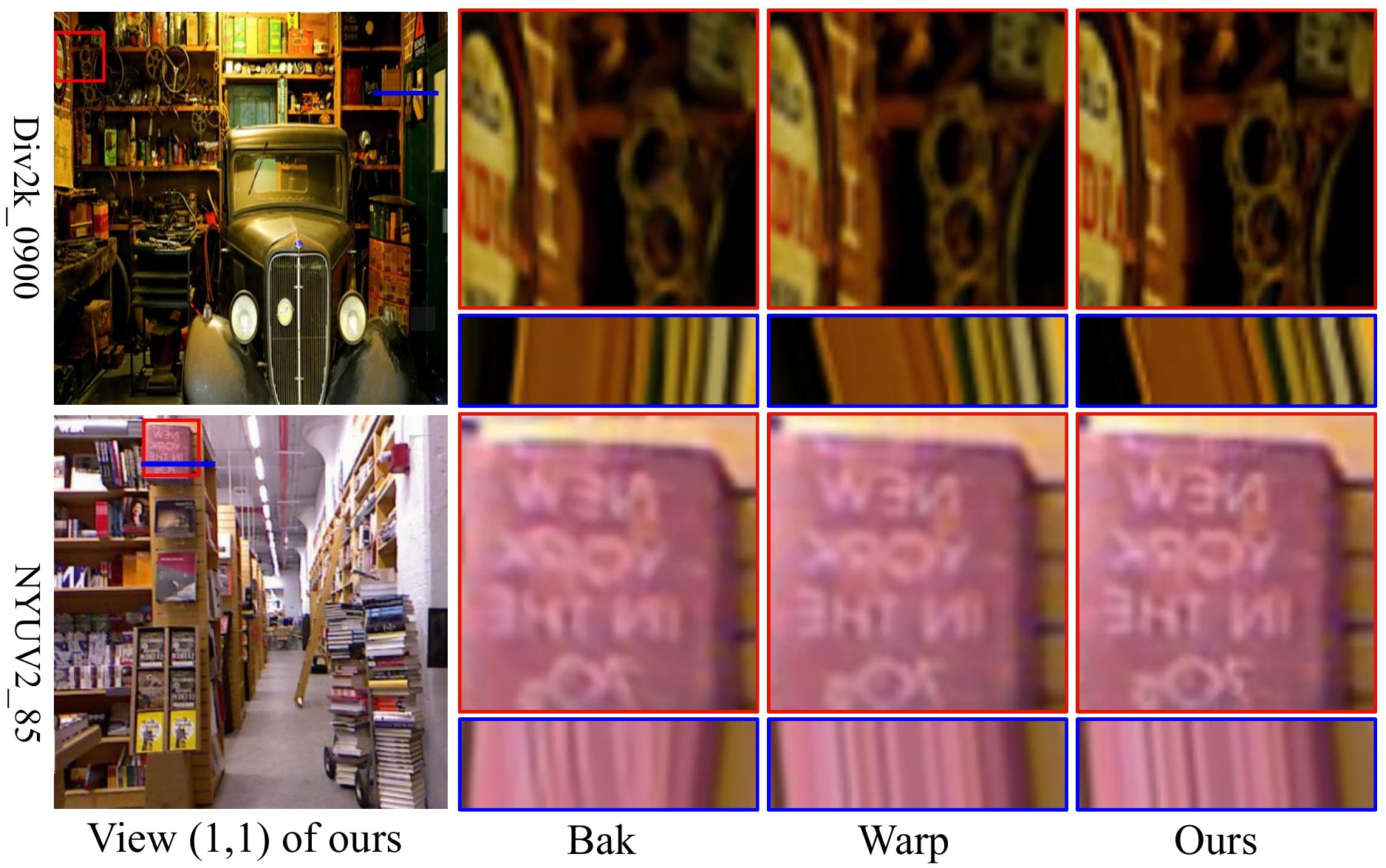}
    \caption{Qualitative comparisons including the SAIs and EPIs of synthesized LFs from single images through different methods (view coordinates: (1, 1)). Zoom in for a better visual experience.}
    \label{Fig:synthesis_visual_2}
\end{figure}

\section{Experiment}
To comprehensively evaluate the effectiveness of LFdiff, the experiments are conducted from two perspectives: (a) LF synthesis from central views and single images on publicly available datasets (Sec.~\ref{sec:4.1} and Sec.~\ref{sec:singleview}) and (b) applications on LF super-resolution and refocusing (Sec.~\ref{sec:4.2} and Sec.~\ref{sec:4.3}). 
We further perform ablation studies in Sec.~\ref{sec:4.4}.

\subsection{LF Synthesis from Central View of LFs} 
\label{sec:4.1}

\noindent\textbf{Baseline Methods and Metrics.} 
We compare LFdiff with four baseline methods, which can be categorized into three classes. 
(a) Warp. We use the estimated disparity to warp the central views based on E.q.~\ref{E.q.warp}. 
(b) Srinivasan \textit{et.al} \cite{srinivasan2017learning} utilizes only central view for input. 
(c) Li \textit{et.al} \cite{li2020synthesizing} and Bak \textit{et.al} \cite{bak2023light} use estimated depths and central views for inputs.  
We adopt PSNR (dB), SSIM~\cite{wang2004image} and LPIPS \cite{zhang2018unreasonable} for evaluating reconstruction fidelity and perceptual quality.

\noindent\textbf{Training Settings.} 
We select 16 scenes from HCI-new \cite{honauer2017dataset} and 170 from UrbanLF-synthetic \cite{sheng2022urbanlf} with angular resolution 5$\times$5 as our training data. During the training stage, LFdiff utilizes the central view and the corresponding ground-truth disparity as input, and each LF is cropped into patches with sizes of $5\times5\times32\times32$. 
We set the total training timesteps $T$ to 1000.
Noise schedule $ \left \{ \beta _{t} \right \} _{t=1}^{T} $ are linearly increased from $\beta _{1}$ = 1e-4 to $\beta _{T}$ = 2e-2. We train the DistgUnet using the AdamW optimizer with batch size 16 and set the initial learning rate to 1.5e-4, scheduled by a cosine annealing scheduler.
We retrain baseline methods with the same training data for a fair comparison.
The training details can be found in the supplementary material.

\noindent\textbf{Inference Settings.}
We select 4 scenes from HCI-new and 30 scenes from UrbanLF-synthetic as our in-distribution testing data and further choose the testset of HCI-old~\cite{wanner2013HCI} and STFGantry~\cite{gantry} to validate the performance in the out-of-distribution testing data.
We utilize a pre-trained monocular depth estimation network \cite{ranftl2020towards} to obtain the normalized invert depth as well as depth for LFdiff and baseline methods, respectively.
The estimated invert depth is further rescaled to disparity which has ground-truth disparity range for LFdiff.
We use DDIM sampler \cite{song2020denoising} with 100 sampling steps for efficient inference.

\noindent\textbf{Quantitative Results.}
We exhibit the quantitative results of our LFdiff and baseline methods in Table~\ref{Tab:syn}.
It can be observed that LFdiff outperforms the existing methods by a large margin in both in-distribution and out-of-distribution testing data. 
For the in-distribution scenes, LFdiff achieves a significant PSNR gain of +1.227dB and +3.846dB compared to the second top-performing method on HCI-new and UrbanLF-synthetic, respectively. 
The clear improvement in LPIPS further shows the superior perceptual quality of our generated results.
Such a performance boost can be attributed to the generation-based pipeline introducing sharper contents and details, especially in occluded regions. 
Furthermore, the position-aware warping condition scheme provides faithful and reliable geometric prior, contributing to our cross-view coherency closer to the ground-truth. 
LFdiff also achieves a performance gain on both fidelity and perceptual metrics for the out-of-distribution scenes.
For example, LFdiff achieves a +1.668dB and +1.924dB performance gain on the metric of PSNR compared to the second top-performing method on HCI-old and STFGantry, respectively.
These results demonstrate that LFdiff exhibits decent generalization capability in the out-of-distribution setting.

\noindent\textbf{Qualitative Results.}
We exhibit the qualitative comparisons corresponding to the top-left view and selected EPIs of LFdiff and baseline methods in Fig.~\ref{Fig:synthesis_visual}. It is evident that the images generated by LFdiff have sharper edges and details, along with fewer artifacts in the occluded regions. For instance, the road sign labeled 'wrong way' generated by LFdiff appears clearer and closer to the ground truth.
As for the edges of the leaves in the HCIold scene, our results exhibit minimal artifacts caused by occlusion. Moreover, LFdiff effectively restores the correct slope direction of HCIold's EPI slice, while also generating the fine details of Urban's EPI slice. This demonstrates the superior performance of LFdiff in terms of angular coherency.

\subsection{LF Synthesis from Single Images}
\label{sec:singleview}
Using the trained models in Sec.~\ref{sec:4.1}, we conduct experiments on LF synthesis from single images in this section by evaluating the performance of LFdiff on two datasets: DIV2K \cite{agustsson2017ntire} and NYUV2 \cite{silberman2012indoor}.
We randomly select a subset of tens scenes for each dataset.
We use two no-reference metrics: NIQE \cite{mittal2012making} as well as BRISQUE \cite{mittal2012no} to evaluate the synthetic performance due to no available ground-truth LFs.
As shown in Table~\ref{Tab:syn2}, LFdiff outperforms other baselines on both testsets, indicating LFs generated by LFdiff contain less distortion and unreal artifacts.
This merit can also be validated in Fig.~\ref{Fig:synthesis_visual_2}.
LFdiff can generate sharper details, such as text and edges while maintaining angular consistency across views compared to other methods.

\subsection{Application: Boosting LF Super-resolution}

\label{sec:4.2}
In this section, we evaluate the data-fulfilling ability of generated LFs by LFdiff.
As a representative downstream task in LF processing, LF super-resolution (LFSR) aims to reconstruct high spatial resolution LFs from low spatial observations with the help of intra-inter view correlations.
Prior works focus on capturing the correlation from multiple LF representations and developing a series of networks \cite{LF-InterNet,cheng2022spatial,xiao2023toward}.
However, from a data perspective, we attempt to improve existing LFSR networks by providing extra training data.
Specifically, we randomly select 160 images from the test split of the NYUV2 dataset and set them as the single image input of LFdiff. 
After estimating the invert depth of these images and cropping them into 32$\times$32 patches, we use LFdiff to create a generated LFSR training set with around 48000 pairs with a disparity range [-3,3], termed NYUV2-LF. 
Following the training and inference setting of the Basic-LFSR framework\footnote{https://github.com/ZhengyuLiang24/BasicLFSR}, we mix the NYUV2-LF with prior LFSR training data and retrain two baseline methods: LF-InterNet \cite{LF-InterNet} and LFSSR-SAV \cite{cheng2022spatial}. 

Table~\ref{Tab:SR} shows the quantitative results of the corresponding methods with or without our extra training data. 
We select Li et.al’s method \cite{li2020synthesizing} (for its best result among baseline methods) to generate the same amount of training data for comparison. 
Benefiting from extra training data, above LFSR baseline methods obtain an apparent performance gain on most metrics, outperforming their original results without any particular design.
Specifically, both methods trained with additional data obtain about 0.4dB performance gain on the STFGantry testset, which shows its enhanced long-range angular information capturing capability. 
Furthermore, extra data provides more unseen details and local structures, improving SSIM performance on all testsets.  

\begin{figure}[tbp!]
    \centering
    \includegraphics[width=0.9\linewidth]{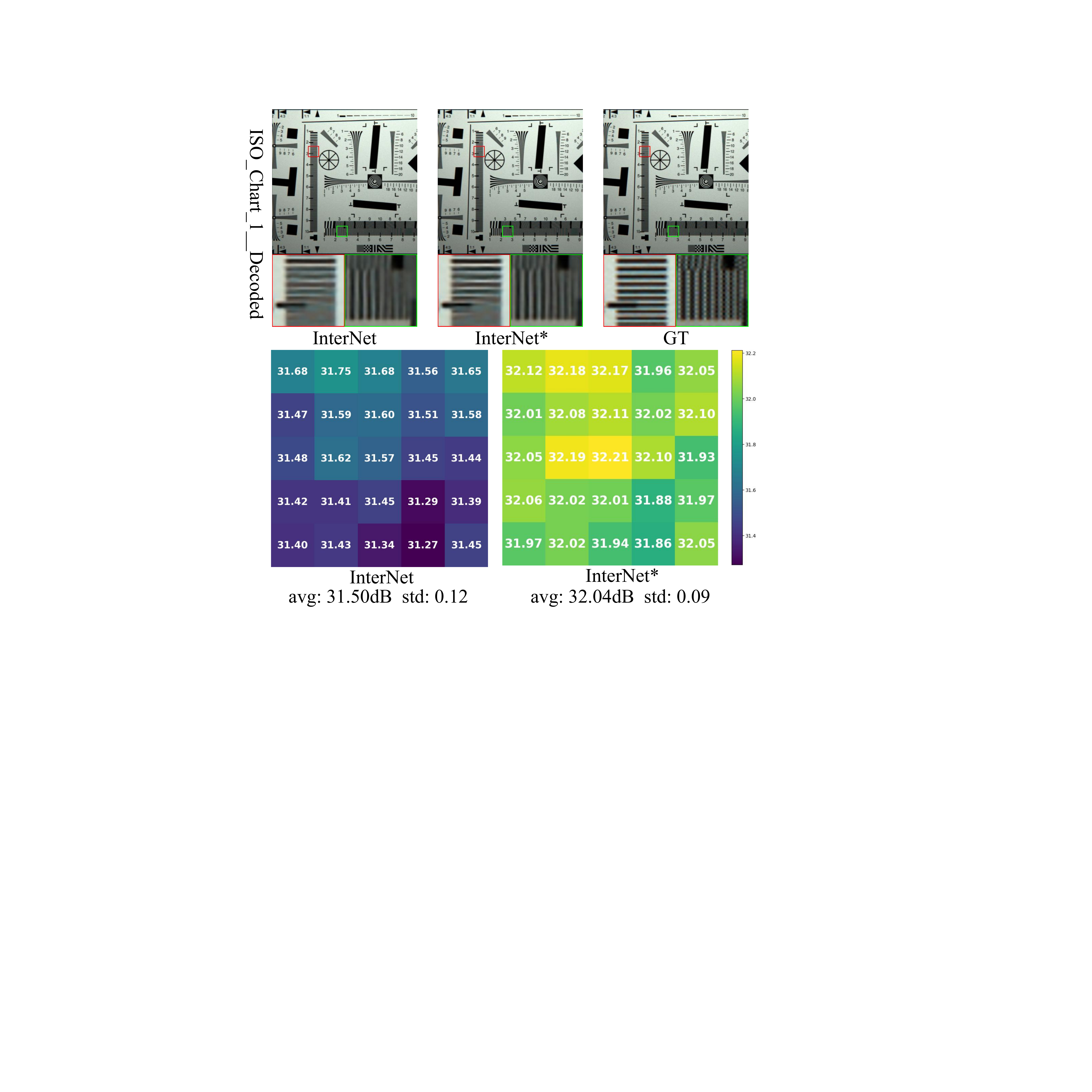}
    \caption{Top: Visual comparison between baseline methods trained w/ or w/o (denoted by *) our generated additional data. Zoom in for a better visual experience. Bottom: PSNR heatmap of the above scene, the model trained with our additional data achieves uniform improvement across views.
    }
    \label{Fig:SR_visual1}
\end{figure}

Fig.~\ref{Fig:SR_visual1} gives a closer look at the qualitative results of different models on the $\times$2 SR task.
We can observe that InterNet cannot handle detailed strip patterns well, whereas InterNet* reconstructs more high-frequency details.
We further show the PSNR heatmaps for view-level improvement comparison in the bottom part of Fig.~\ref{Fig:SR_visual1}.
Models trained with additional data achieve a uniform gain (i.e. less std value) among views, validating the benefits of angular information provided by our generated data.

\begin{figure}[ht!]
    \centering
    \includegraphics[width=0.995\linewidth]
    {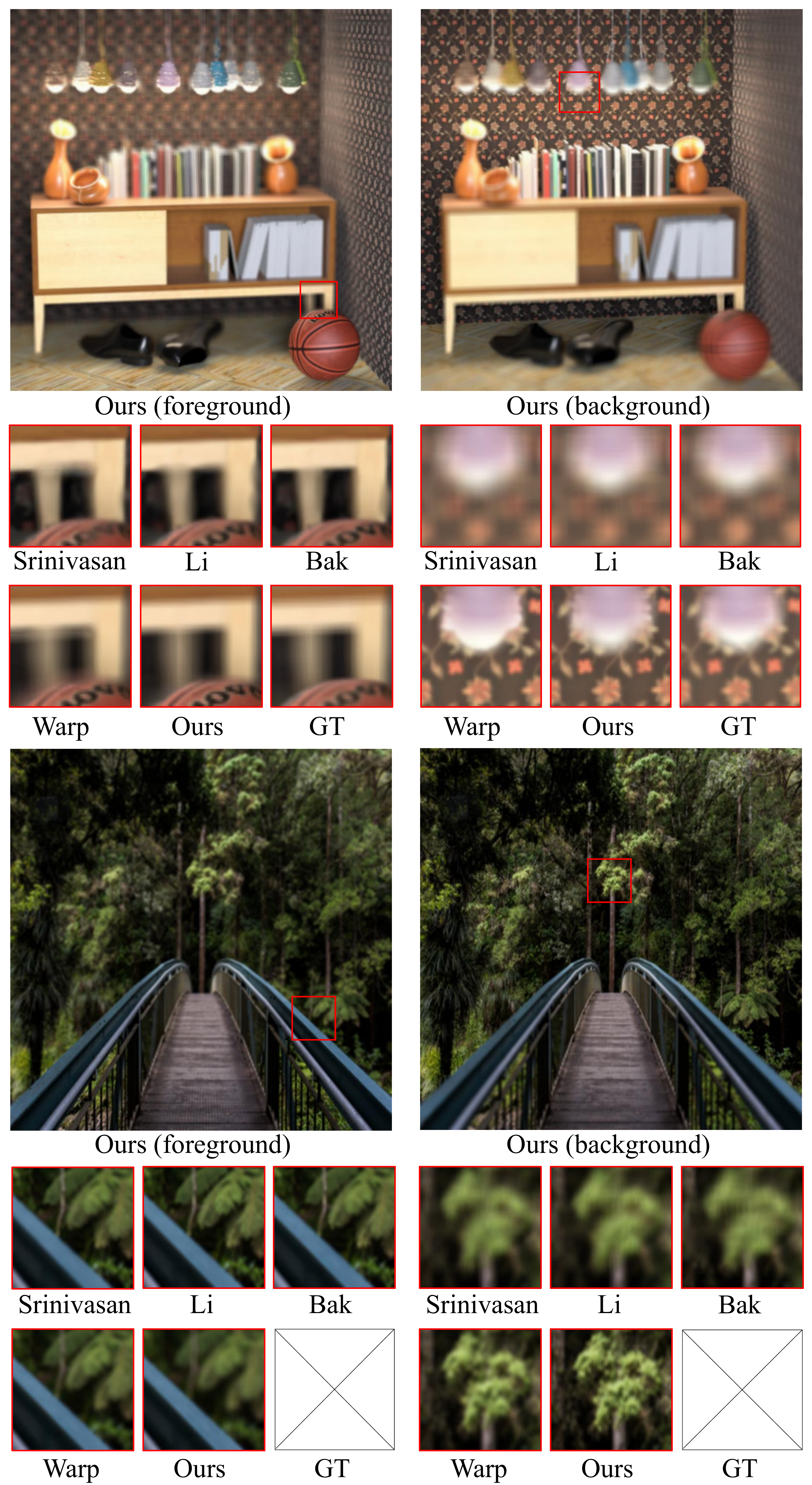}
    \caption{Refocus results on scenes from HCI-new-sideboard and DIV2K-0876. Zoom in for a better visual experience.}
    \label{Fig:refocus_visual}
\end{figure}

\begin{table*}[ht!]
\centering
\caption{Quantitative results (PSNR / SSIM) on the $\times$2 LF super-resolution task with the input angular resolution of 5$\times$5. Networks trained with generated addition data by Li \textit{et.al} and ours are denoted as $\dagger$ and *, respectively. We mark the better results in \textbf{bold}.}
\label{Tab:SR}
\scalebox{0.85}{
\begin{tabular}{@{}llllll|l@{}}
\toprule
Method & \multicolumn{1}{c}{HCI-new} & \multicolumn{1}{c}{HCI-old} & \multicolumn{1}{c}{EPFL} & \multicolumn{1}{c}{INRIA} & \multicolumn{1}{c}{STFGantry} & \multicolumn{1}{c}{Average}\\ \midrule
LF-InterNet \cite{LF-InterNet}  & 37.170/0.9529 &  44.573/0.9875 &  34.112/0.9584 &  35.829/0.9655 &  38.435/0.9852 & 38.024/0.9699 \\
LF-InterNet$\dagger$  & 37.118/0.9528 &  44.483/0.9873 &  34.333/0.9584 &  36.163/0.9655 & 38.380/0.9855 & 38.095/0.9699 \\
LF-InterNet* &  \textbf{37.251/0.9536}	& \textbf{44.577/0.9876}	& \textbf{34.366/0.9588} & \textbf{36.227/0.9658} & \textbf{38.820/0.9874} & \textbf{38.248/0.9707} \\ \midrule
LFSSR-SAV \cite{cheng2022spatial} &   \textbf{37.425}/0.9556   &  44.215/0.9866 & \textbf{34.616}/0.9600 & 36.364/0.9664 & 38.689/0.9861 & 38.262/0.9710 \\
LFSSR-SAV$\dagger$  & 37.348/0.9551 &  44.300/0.9870 &  34.585/0.9603 &  36.377/0.9667 &  38.837/0.9867 & 38.289/0.9712 \\
LFSSR-SAV* &  37.398/\textbf{0.9558}  & \textbf{44.370/0.9871} & 34.597/\textbf{0.9604} & \textbf{36.389/0.9668} & \textbf{39.068/0.9878} & \textbf{38.364/0.9716} \\ \bottomrule
\end{tabular}
}
\end{table*}

\subsection{Application: Refocusing}
\label{sec:4.3}
Compared to conventional 2D photography, LF imaging provides opportunities for post-capture refocusing.
The extra angular information allows the post-exposure alternation of focal planes via the integral transform \cite{ng2005fourier}.
As shown in Fig.~\ref{Fig:refocus_visual}, we provide the refocus results of the same foreground-background position on HCI-new and DIV2K scenes for a fair comparison.
LFs produced by our method achieve correct sharp/blurry effects compared to other baselines.
For example, when refocusing on the background in the HCI-new scene, the foreground purple light and background flower patterns are supposed to be blurry and sharp, respectively.
Our results are the only ones that satisfy this requirement.
Our method also performs well in the single image input scenario.
For example, when focusing on the foreground bridge, the leaves in the background are supposed to be blurry.
When focusing on the background trees, the detail within the tree region needs to be clear.
Our method shows a clear advantage in both situations and has the most distinctive visual quality in depth-variant regions, which is shown in the bottom part of Fig.~\ref{Fig:refocus_visual}.

\subsection{Ablation Studies}
\label{sec:4.4}
In this section, we conduct ablation studies on the conditional mechanism and the effect of disparity range. 
More results can be found in the supplementary material.

\begin{figure}[ht!]
    \centering
    \includegraphics[width=0.9\linewidth]
    {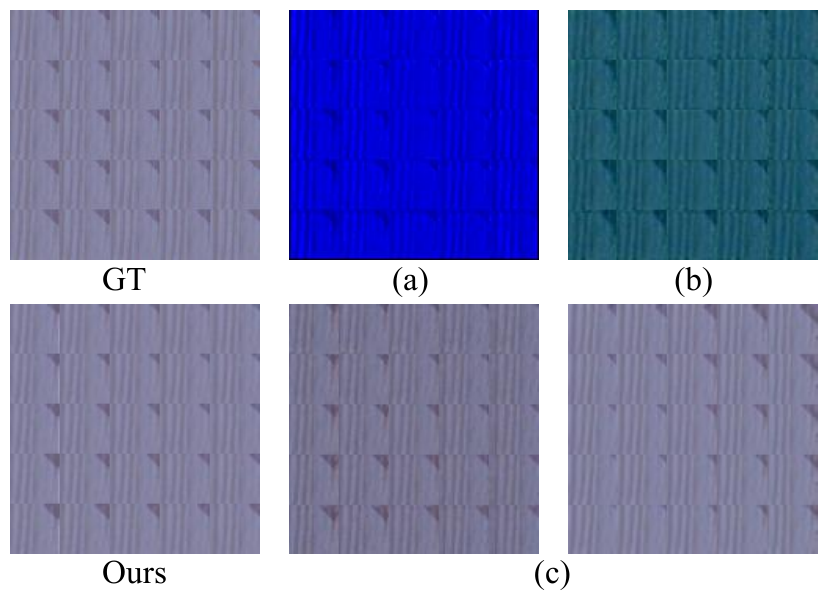}
    \caption{Different condition methods result in different generation results. We use a generated patch for better comparison. (a) central view only. (b) Disparity embedding. (c) Warping without position encoding. }
    \label{Fig:ablation}
\end{figure}

\noindent \textbf{Condition Mechanism.}
We conduct ablation studies on HCI-new to evaluate the effectiveness of the position-aware warping condition scheme. 
The baseline methods are set based on the different degrees of participation of angular information: (a) no disparity, (b) implicit disparity guidance and (c) explicit disparity guidance without position encoding.
Given the same noise estimation networks, we construct these baselines with different condition methods: (a) central view only, (b) disparity embedding and (c) warp without position encoding. 
The visual results are shown in Fig.~\ref{Fig:ablation}.
As for (a) and (b), the model tends to directly learn the complex distribution of spatial-angular SAIs without explicit angular information, resulting in generating spatial-degraded samples with severe color shifting.
Although we obtain reasonable results when using (c), there may occasionally arise scenarios where the disparity condition fails to exert complete control over the output geometry, leading to inverse angular patterns. The aforementioned phenomenon can be mitigated by utilizing a view-level position encoding in our solution.

\begin{figure}[ht!]
    \centering
    \includegraphics[width=\linewidth]
    {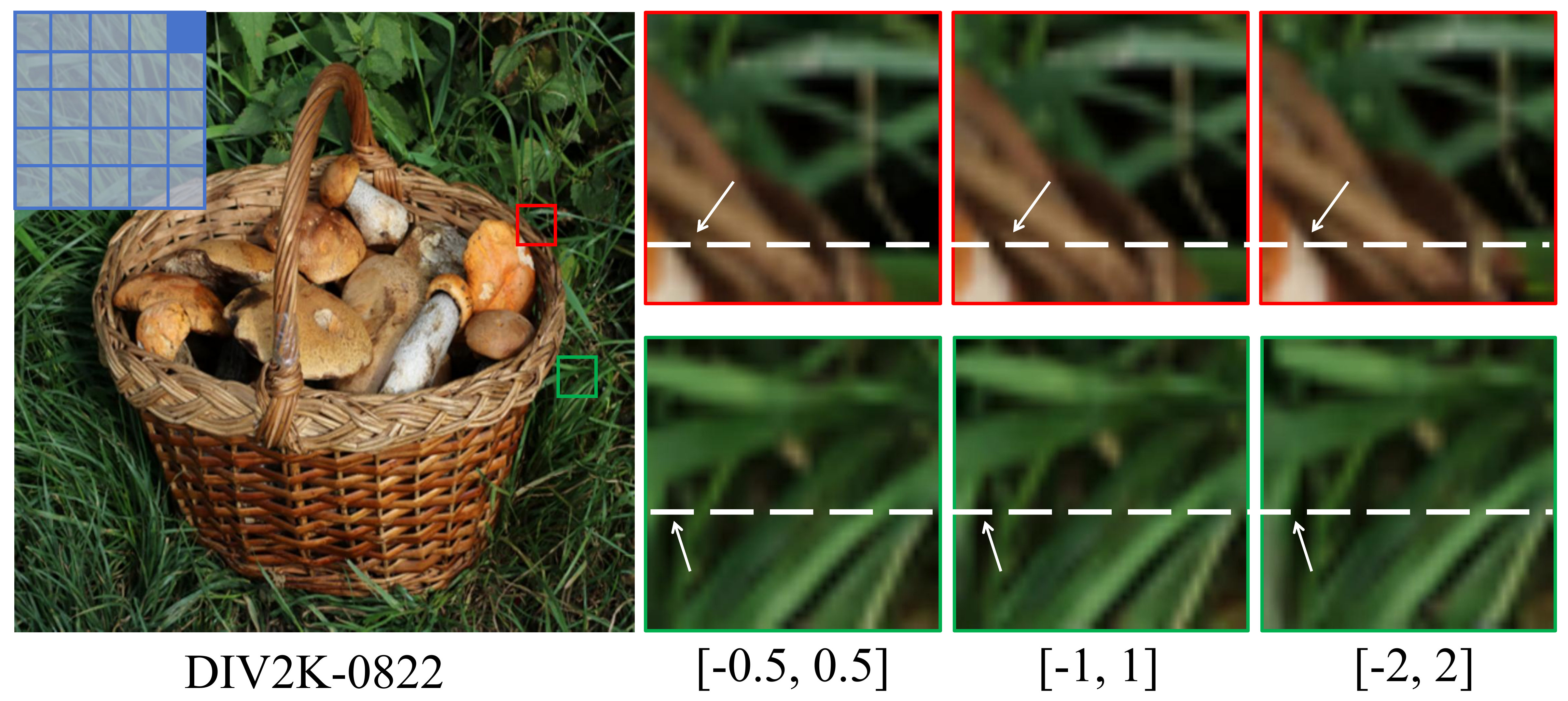}
    \caption{Our method is aware of the input disparity range. We rescale the estimated disparity to three different ranges([-0.5,0.5], [-1,1] and [-2,2]) and generate the corresponding corner view. White lines and arrows are provided for clarity.}
    \label{Fig:ablation2}
\end{figure}

\noindent \textbf{Disparity Range.} Benefiting from explicit disparity control, LFdiff allows for controlled generation under different disparity ranges. 
Here, we provide some visual results on how different disparity range affects the generation results.
We select a single image from DIV2K and estimate its normalized invert depth, which is then rescaled to disparity with three different ranges: [-0.5, 0.5], [-1, 1] and [-2, 2].
The generated results using different disparities are shown in Fig.~\ref{Fig:ablation2}. We select a corner view for illustration.
With the increase of the disparity range, LFdiff generates varying details corresponding to the disparity value in the same spatial-angular location.
For example, the gradual revealing of the left lower corner in the red patch and the upside leaf in the green patch shows the impact of disparity range controls.

\section{Conclusion}

We introduce LFdiff, the first conditional diffusion framework tailored for generating LFs from single RGB images.
Specifically, we propose a position-aware condition scheme to generate an initial LF condition, which drives the diffusion model to learn the robust inter-view geometry.
We then incorporate the disentangle mechanism into the noise estimation network to leverage comprehensive LF representations from macro-pixel form inputs.
Extensive experiments show that LFDiff achieves superior quantitative and qualitative results both in in-distribution and out-of-distribution scenarios.
The generated LFs from our framework demonstrate their effectiveness in various downstream applications, \textit{e.g.}, LF super-resolution and refocusing.

{
    \small
\bibliographystyle{ieeenat_fullname}
    \bibliography{main}
}

\end{document}